\documentclass{article}

\usepackage{microtype}
\usepackage{graphicx}
\usepackage{subfigure}
\usepackage{booktabs} 
\usepackage{amsmath}
\usepackage{amssymb}
\usepackage{mathtools}
\usepackage{filecontents}
\usepackage{pgfplots}
\usepgfplotslibrary{fillbetween}
\usetikzlibrary{arrows}
\usetikzlibrary{backgrounds}
\usepackage{xcolor}

\usepackage{hyperref}

\usepackage[accepted]{icml2020}

\icmltitlerunning{Graph Convolutional Gaussian Processes for Link Prediction}

\newcommand{\bd}[1]{\mathbf{#1}}
\newcommand{\R}{\mathbb{R}}
\newcommand{\LO}{\mathcal{O}}
\newcommand{\GP}{\text{GP}}
\newcommand{\normal}{\mathcal{N}}
\newcommand{\expect}{\mathbb{E}}
\newcommand{\KL}[2]{\text{KL}[#1 \;\|\; #2]}
\definecolor{highlightBlue}{RGB}{58,146,222}
\definecolor{highlightOrange}{RGB}{243,165,54}

\begin{document}

\twocolumn[
\icmltitle{Graph Convolutional Gaussian Processes for Link Prediction}




\begin{icmlauthorlist}
\icmlauthor{Felix L. Opolka}{cam}
\icmlauthor{Pietro Li\`o}{cam}
\end{icmlauthorlist}

\icmlaffiliation{cam}{Department of Computer Science and Technology, University of Cambridge, Cambridge, United Kingdom}

\icmlcorrespondingauthor{Felix Opolka}{flo23@cam.ac.uk}

\icmlkeywords{Machine Learning, ICML, Gaussian Process, Graphs}

\vskip 0.3in
]



\printAffiliationsAndNotice{\icmlEqualContribution} 

\begin{abstract}
Link prediction aims to reveal missing edges in a graph. We address this task with a Gaussian process that is transformed using simplified graph convolutions to better leverage the inductive bias of the domain. 
To scale the Gaussian process model to large graphs, we introduce a variational inducing point method that places pseudo inputs on a graph-structured domain. 
We evaluate our model on eight large graphs with up to thousands of nodes and report consistent improvements over existing Gaussian process models as well as competitive performance when compared to state-of-the-art graph neural network approaches.
\end{abstract}

\section{Introduction}

A large variety of real-world scenarios can be modelled by signals that live on the nodes of a graph: from biological networks to communication and social networks~\cite{sen2008collective, kersting2016benchmark}.
The connective structure of these graphs is not necessarily complete, hence a common task for statistical inference is to infer missing links between nodes~\cite{wang2015lpsocial}.
In a protein-protein interaction network, for example, link prediction is used to suggest interactions between two proteins~\cite{lei2012ppi}.

Recent work in this area~\cite{kipf2016gvae, zhang2018link} has focused on methods with two key properties.
First, these methods can predict missing links based on both the graph structure itself and a signal that lives on the nodes of the graph, often referred to as the node features.
Second, these methods compute node embeddings not only from isolated features of each node but also take into account features in the local neighbourhood of each node, thus providing more context information for predicting missing links.
At the core of these methods are usually neural networks equipped with parameterised graph convolution operations.

While these neural network models achieve state-of-the-art performance~\cite{zhang2018link}, they require considerable amounts of labelled data due to a large number of parameters that are optimised using maximum likelihood estimation.
In this work, we propose to address the link prediction task with a Gaussian process model.
Crucially, the proposed model is aware of both the graph structure and the node features and makes use of graph convolutions to incorporate neighbourhood information when predicting links.
It naturally counters overfitting by marginalising parameters in a Bayesian inference setting and optimising hyperparameters using a variational lower bound, thus requiring no validation set for early stopping.
Furthermore, the Gaussian process model provides a principled way of obtaining uncertainty estimates, which are often required for downstream tasks.

We derive the graph convolutional Gaussian process for link prediction in two steps.
First, we introduce a versatile graph convolutional Gaussian process model over the \textit{nodes} of a graph.
This is achieved by transforming a Gaussian process defined on the Euclidean domain with graph convolutions.
Second, we further adapt the resulting model over nodes to a Gaussian process over \textit{pairs of nodes}, thus suitable for link prediction.
We present a scalable variational approximation of the posterior distribution of the model with inducing points.
In summary, our paper makes the following contributions:
    \begin{itemize}
    \item We introduce a new graph convolutional Gaussian process model operating on the nodes of a graph. It is trained to automatically fit the neighbourhood size of the graph convolutions to the input graph (Section~\ref{sec:gcgp}).
    \item We present the first graph convolutional variational Gaussian process model for link prediction (Section~\ref{sec:link_pred}).
    \item We suggest a variational inducing point method for link prediction that works by placing inducing points on the nodes of an inducing graph (Section~\ref{sec:link_pred}).
    \item We evaluate our method on a range of benchmark data sets with several thousand nodes and edges and achieve competitive performance compared to other link prediction methods, such as the variational graph auto-encoder~\cite{kipf2016gvae} (Section~\ref{sec:experiments}).
    \end{itemize}

\section{Background}

\subsection{Gaussian Processes}

A Gaussian process models functions as samples from an infinite dimensional multivariate normal distribution.
The shape of the functions are determined by the mean and covariance (or kernel) function of the process.
When modelling observed data $\mathcal{D}= (\bd{X}, \bd{y})$ with input data matrix $\bd{X} = [\bd{x}_1, \ldots, \bd{x}_N]^T$, $\bd{x}_i \in \mathcal{X}$ and labels $\bd{y} \in \R^{N}$ via Bayesian inference, we can use a Gaussian process as the prior distribution over the latent function:
    \begin{equation}
        f(\bd{x}) \sim \GP(m(\bd{x}), k_{\theta}(\bd{x}, \bd{x'})),
    \end{equation}
where $m: \mathcal{X} \rightarrow \R$ and $k_\theta: \mathcal{X} \times \mathcal{X} \rightarrow \R$ denote the mean and covariance function respectively.
The covariance function $k_\theta$ is commonly parameterised by a set of hyperparameters $\theta$.

When combined with a Gaussian likelihood $p(y_n \vert \bd{x}_n)$ for each observation $n = 1, \ldots, N$, the posterior $p(\bd{f} \vert \bd{y}, \bd{X})$ is also Gaussian.
Predictions for new data points can then be made in a fully Bayesian fashion by marginalising out the latent function $f(\bd{x})$. 
Furthermore, the marginal likelihood $p(\bd{y})$ has a closed form solution and can thus be used to optimise the kernel hyperparameters $\theta$, usually via gradient-based optimisation.
For our purposes, we set $\mathcal{X} = \R^{D}$, hence $\bd{X} \in \R^{N\times D}$.

This formulation of Gaussian processes is limited in two ways.
Firstly, when the likelihood is not Gaussian, as is the case for link prediction, neither the posterior distribution nor the marginal likelihood have a closed-form solution.
Secondly, inference with a Gaussian process requires the inversion of an $N \times N$ matrix, which has complexity $\LO(N^3)$ and is thus infeasible for large data sets.
Both problems are commonly addressed by approximating the intractable posterior with a variational posterior distribution evaluated at a small set of \textit{inducing points} $\bd{Z} = [\bd{z}_1, \ldots, \bd{z}_M]^T$, with $\bd{z}_i \in \mathcal{X}$ and $M \ll N$.
The inducing points are assumed to follow the same Gaussian process prior distribution as the original inputs, hence the inducing points give rise to a set of inducing variables $\bd{u} = [f(\bd{z}_1), \ldots, f(\bd{z}_M)]^T$.
Accordingly, the inducing points follow the prior distribution $p(\bd{u}) = \normal(\bd{m}_\bd{z}, \bd{K}_{\bd{z}\bd{z}})$, where $[\bd{m}_\bd{z}]_i = m(\bd{z}_i)$ and $[\bd{K}_{\bd{z}\bd{z}}]_{ij} = k_\theta(\bd{z}_i, \bd{z}_j)$.
Inference for a new input $\bd{x^*}$ is now performed using the \textit{sparse} Gaussian process over the inducing points:
    \begin{align}
        f(\bd{x^*}) \vert \bd{u} \sim \GP(&\bd{k}_{\bd{zx^*}}^T\bd{K}_{\bd{zz}}^{-1}\bd{u},\\
        &k_\theta(\bd{x^*}, \bd{x^*}) - \bd{k}_{\bd{zx^*}}^T\bd{K}_{\bd{zz}}^{-1}\bd{k}_{\bd{zx^*}}),
    \end{align}
where $\bd{k}_{\bd{zx^*}} = [k_\theta(\bd{z}_1, \bd{x^*}), \ldots, k_\theta(\bd{z}_M, \bd{x^*})]$.
The inducing points can be considered a compressed version of the original input data set.

The variational distribution is chosen to be a multivariate Gaussian distribution $q(\bd{u}) = \normal(\bd{m}, \bd{S})$.
The inducing points $\bd{Z}$, as well as $\bd{m}$ and $\bd{S}$ are variational parameters, which are optimised jointly with the kernel hyperparameters $\theta$ by maximizing the Evidence Lower Bound (ELBO) objective:
    \begin{align}
        \mathcal{L}(\theta, \bd{Z}, \bd{m}, \bd{S}) = &\sum_{n=1}^{N} \expect_{q(f(\bd{x}_n)}[\log p(y_n \vert f(\bd{x}_n))] \nonumber\\
        &- \KL{q(\bd{u})}{p(\bd{u})}.
    \end{align}
The shape of this objective enables optimisation via stochastic gradient descent, which reduces the memory complexity of an individual update step, thus allowing us to train on larger data sets.

\subsection{Graph Convolutions}


The graph convolutional neural network~\cite{kipf2017gcn} is one of the most widely used graph neural networks, shown to achieve good performance on various tasks (see for example~\citealt{shchur2019pitfalls}).
It aims to seize the inductive bias of the domain by specifically encoding localised patterns inside node representations. This is achieved by taking inspiration from convolutional neural networks for images~\cite{lecun1999cnn} and generalising the convolution operation from the domain of regular grids to the domain of general graphs~\cite{bruna2014graphcnn}. The graph convolution is applied to an input signal $\bd{X} \in \R^{N \times D}$ lying on the domain of a graph $\mathcal{G} = (\mathcal{V}, \mathcal{E})$ with node set $\mathcal{V}$, edge set $\mathcal{E}$, and adjacency matrix $\bd{A} \in \R^{N \times N}$ (without self-loops).
The convolution operator is formulated as a multiplication of the filter with the input signal mapped to the spectral domain via the Fourier transform. Analogously to the Fourier transform on the Euclidean domain, the graph Fourier transform is defined as the decomposition of a signal into the eigenfunctions of the Laplace operator. On the graph domain, this operator is given by the Laplace matrix $\bd{L} = \bd{A} - \bd{D}$, where $\bd{D} \in \R^{N\times N}$ is the diagonal degree matrix with $D_{ii} = \sum_{j=1}^{N} A_{ij}$.
To localise the convolution operation, the filter is commonly parameterised with Chebyshev polynomials in the spectral domain, as proposed by~\citealt{defferrard2016chebnet}.


\section{Related Work}

Our work is closely related to the two fields of Gaussian processes for graph-structured data and link prediction.

\paragraph{Gaussian processes for graph-structured data}

Prior Gaussian process models for graph-structured data have been studied under the term relational learning. These methods have been applied to semi-supervised classification of nodes in a graph, such as the relational Gaussian process~\cite{chu2007rgp}, the mixed graph Gaussian process~\cite{silva2008xpr}, or the label propagation algorithm~\cite{zhu2003lp, zhu2003lpgp}.
Inspired by graph neural networks, more recent work has developed Gaussian process models that explicitly consider nodes together with node features in their local neighbourhood.
The graph Gaussian process described by~\citealt{ng2018bayesian} computes node representations by averaging the node features of the 1-hop neighbourhood and subsequently performing semi-supervised node classification. Unlike the graph convolutional Gaussian process proposed here, it only considers 1-hop node neighbourhoods, thus limiting the node neighbourhood information accessible to the model.
The graph convolutional Gaussian process introduced by~\citealt{wilk2017cgp} employs graph convolutions to produce representations of patches in the graph and sums up these patches via an additive Gaussian process model. Unlike the models described so far, it is used for graph-level prediction such as image or mesh classification.
To the best of our knowledge, the only other Gaussian process model for link prediction is described by~\citealt{yu2008link}. However, it does not include information from node neighbourhoods, which restricts its predictive performance.

\paragraph{Link prediction models}

A common class of link prediction methods is represented by heuristic-based models, explored systematically by~\citealt{zhang2018link}.
These methods compute heuristics for node similarity and output it as the likelihood of a link. Popular heuristics include common neighbours, Jaccard, preferential attachment~\cite{barabasi1999pa}, Adamic-Adar~\cite{adamic2001aa}, resource allocation~\cite{zhou2009ra}, Katz, rooted PageRank~\cite{brin2012google}, and SimRank~\cite{jeh2002simrank}.

Other methods focus on predicting links based on latent node features that are derived from the graph structure. For example, the node features computed via spectral clustering can be used for link prediction. Other latent feature methods are matrix factorisation (MF)~\cite{koren2009matrixfact} and the stochastic block model (SBM) for link prediction~\cite{airoldi2008block}.
More recent approaches such as DeepWalk~\cite{perozzi2014deepwalk}, LINE~\cite{tang2015line}, and node2vec~\cite{grover2016node2vec} rely on random walks to produce node embeddings that encode latent features and pair-wise comparison of the embeddings to predict links. These approaches can also be cast as matrix factorisation~\cite{qiu2018unifying}.
Naturally, matrix factorisation methods do not consider node features.

Another class of link prediction methods makes use of neural networks. The Weisfeiler-Lehman Neural Machine (MLNM)~\cite{zhang2017wlnm} trains a fully-connected neural network on adjacency matrices.
SEAL~\cite{zhang2018link} employs graph-neural networks in a non-probabilistic setting. The network operates on node features and hand-crafted node labels that indicate a node's role in its neighbourhood.
Most similar to our model, the graph variational auto-encoder by~\citealt{kipf2016gvae} combines probabilistic modelling and graph convolutions, thus also considering neighbourhood information.
It samples node embeddings $\bd{z}_i$ from a normal distribution
    \begin{equation}
        q(\bd{z}_i \vert \bd{X}, \bd{A}) = \normal(\bd{z}_i \vert \bd{\mu}, \bd{\Sigma}),
    \end{equation}
where the mean $\bd{\mu}$ and variance $\bd{\Sigma}$ are computed by a graph convolutional neural network~\cite{kipf2017gcn}. It independently predicts links using the generative model
    \begin{equation}
        p(A_{ij} = 1 \vert \bd{z}_i, \bd{z}_j) = \sigma(\bd{z}_i^T\bd{z}_j).
    \end{equation}
The form of the distribution over links is more restricted due to its reliance on the inner product between node representations. In contrast, we achieve high flexibility by choosing the variational distribution to be a Gaussian process evaluated at inducing points, which are free parameters themselves.

The Gaussian process model proposed in the following sections exhibits many of the individual strong points of existing models. It considers both graph structure and node features and incorporates local neighbourhood information when inferring missing links. Moreover, the Bayesian inference framework provides us with a principled way of obtaining uncertainty estimates for our predictions.

\section{Graph Convolutional Gaussian Processes}\label{sec:gcgp}

As a first step, we describe a graph convolutional Gaussian process over nodes before adapting it to the task of link prediction in Section~\ref{sec:link_pred}. We aim to define a Gaussian process model that is capable of seizing the inductive bias of the domain whose structure is given by an undirected graph $\mathcal{G} = (\mathcal{V}, \mathcal{E})$ with a set of vertices $\mathcal{V}$, $\vert \mathcal{V} \vert = N$, and a set of edges $\mathcal{E}$, $\vert \mathcal{E} \vert = E$.
The graph structure is further described by the adjacency matrix $\bd{A}$ without self-loops, i.e. its diagonal entries are $0$.
Input data is given in form of a signal $\bd{X} \in \R^{N \times D}$ living on said domain.

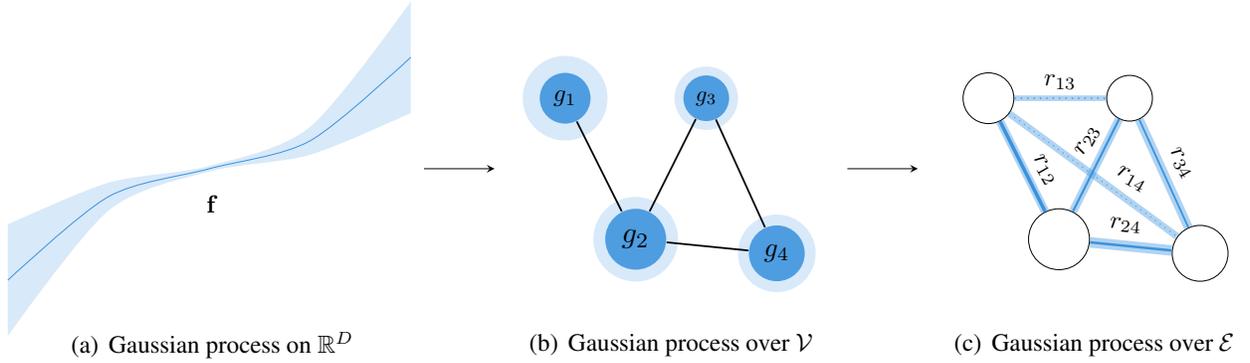
\begin{figure*}
    \centering
    \newsavebox\mysubpic
\sbox{\mysubpic}{%
    \begin{tikzpicture}[remember picture]
        \begin{axis}[hide axis]
        \addplot[smooth,draw=highlightBlue] table[x=x,y=y] {data.dat};
        \addplot[name path=upper,draw=none,smooth] table[x=x,y expr=\thisrow{y}+\thisrow{err}] {data.dat};
        \addplot [name path=lower,draw=none,smooth] table[x=x,y expr=\thisrow{y}-\thisrow{err}] {data.dat};
        \addplot [fill=highlightBlue!20] fill between[of=upper and lower];
        \end{axis}
    \end{tikzpicture}
}

\resizebox{\textwidth}{!}{%
\begin{tikzpicture}

\node (a) at (0.0, 0.0) {\usebox{\mysubpic}};
\node at (0.0, -0.5) {$\bd{f}$};
\node at (0.0, -2.5) {(a)\, Gaussian process on $\R^D$};

\draw[->, >=stealth] (3.0, 0.0) -- (4.0, 0.0);

\fill[fill=highlightBlue!20] (5.0, 1.0) circle[radius=0.6];
\node[circle, scale=1.0, color=black, fill=highlightBlue!90] (g1) at (5.0, 1.0) {$g_1$};
\fill[fill=highlightBlue!20] (6.0, -1.0) circle[radius=0.6];
\node[circle, scale=1.2, color=black, fill=highlightBlue!90] (g2) at (6.0, -1.0) {$g_2$};
\fill[fill=highlightBlue!20] (7.0, 1.0) circle[radius=0.45];
\node[circle, scale=0.9, color=black, fill=highlightBlue!90] (g3) at (7.0, 1.0) {$g_3$};
\fill[fill=highlightBlue!20] (8.0, -1.2) circle[radius=0.55];
\node[circle, scale=1.1, color=black, fill=highlightBlue!90] (g4) at (8.0, -1.2) {$g_4$};
\draw[line width=0.25mm] (g1) -- (g2);
\draw[line width=0.25mm] (g2) -- (g3);
\draw[line width=0.25mm] (g2) -- (g4);
\draw[line width=0.25mm] (g3) -- (g4);
\node at (6.5, -2.5) {(b)\, Gaussian process over $\mathcal{V}$};

\draw[->, >=stealth] (9.0, 0.0) -- (10.0, 0.0);

\node[circle, scale=1.0, color=white, fill=white, draw=black] (g1) at (11.0, 1.0) {$g_1$};
\node[circle, scale=1.2, color=white, fill=white, draw=black] (g2) at (12.0, -1.0) {$g_2$};
\node[circle, scale=0.9, color=white, fill=white, draw=black] (g3) at (13.0, 1.0) {$g_3$};
\node[circle, scale=1.1, color=white, fill=white, draw=black] (g4) at (14.0, -1.2) {$g_4$};
\draw[draw=highlightBlue, draw opacity=0.4, line width=1.2mm] (g1) -- (g2);
\draw[draw=highlightBlue, line width=0.5mm] (g1) -- node[above, rotate=-60] {\hspace{0.4cm}$r_{12}$} (g2);
\draw[draw=highlightBlue, draw opacity=0.4, line width=0.8mm] (g1) -- (g3);
\draw[draw=highlightBlue, line width=0.2mm, dotted] (g1) -- node[above] {$r_{13}$} (g3);
\draw[draw=highlightBlue, draw opacity=0.4, line width=1.0mm] (g1) -- (g4);
\draw[draw=highlightBlue, line width=0.15mm, dotted] (g1) -- node[above, rotate=-30] {\hspace{1.0cm}$r_{14}$} (g4);
\draw[draw=highlightBlue, draw opacity=0.4, line width=1.2mm] (g2) -- node[above, rotate=60] {\hspace{0.5cm}$r_{23}$} (g3);
\draw[draw=highlightBlue, line width=0.3mm] (g2) -- (g3);
\draw[draw=highlightBlue, draw opacity=0.4, line width=1.5mm] (g2) -- node[above, rotate=-5] {\hspace{-0.2cm}$r_{24}$} (g4);
\draw[draw=highlightBlue, line width=0.35mm] (g2) -- (g4);
\draw[draw=highlightBlue, draw opacity=0.4, line width=1.3mm] (g3) -- node[above, rotate=-60] {$r_{34}$} (g4);
\draw[draw=highlightBlue, line width=0.25mm] (g3) -- (g4);
\node at (12.5, -2.5) {(c)\, Gaussian process over $\mathcal{E}$};

\end{tikzpicture}
}
    \caption{Overview of the proposed graph convolutional Gaussian process model for link prediction. We start with a regular Gaussian process $\bd{f}$ (a) operating solely on the node features that is oblivious to the graph structure. Each node feature is treated as an observation on the Euclidean domain $\R^{D}$. This Gaussian process is transformed using simplified graph convolutions to yield a graph convolutional Gaussian process $\bd{g}$ over the nodes $\mathcal{V}$ of the graph (b). Finally, a series of such graph convolutional Gaussian processes as defined in Equation~\ref{eq:edge_func} yields a graph convolutional Gaussian process $\bd{r}$ over edges (c). Function values in (b) and (c) are expressed through the size of the nodes and the thickness of the links respectively. Confidence intervals are sketched in light blue.}
    \label{fig:model_summary}
\end{figure*}

\subsection{Model Formulation}

In a non-probabilistic setting, adaption to the graph domain is commonly achieved through graph convolutional neural network layers formulated by~\citealt{kipf2017gcn}. Crucially, these augment the multi-layer perceptron (MLP) with an aggregation step carried out directly after the linear mapping of the node features and before the non-linearity:
    \begin{equation}
        \bd{\bar{h}}_i = \sum_{j \in \mathcal{N}(i) \cup \{i \}} \frac{1}{\sqrt{(d_i + 1)(d_j + 1)}} \bd{h}_j,\label{eq:agg_rule}
    \end{equation}
where $\bd{h}_j$ are the node features after the linear map, $\mathcal{N}(i)$ denotes the 1-hop neighbourhood around node $i$, and $d_j$ denotes the degree of node $j$.
This aggregation step can be expressed as a multiplication of the node feature matrix with the normalised adjacency matrix $\bd{\tilde{S}} = \bd{\tilde{D}}^{-\frac{1}{2}} \bd{\tilde{A}} \bd{\tilde{D}}^{-\frac{1}{2}}$:
    \begin{equation}
        \bar{\bd{H}} = \bd{\tilde{S}}\bd{H},
    \end{equation}
where $\bd{\tilde{A}} = \bd{A} + \bd{I}$ is the adjacency matrix with added self-loops and $\bd{\tilde{D}}$ is the degree matrix of $\bd{\tilde{A}}$.
The combination of the linear mapping followed by the aggregation step is referred to as a graph convolution and allows the neural network to produce embeddings that capture local features in the neighbourhood of a node.

While~\citealt{kipf2017gcn} propose to stack layers each consisting of a linear map, followed by the agreggation step and a non-linearity,~\citealt{wu2019sgcn} have shown that equivalent performance can be achieved through \textit{simplified graph convolutions}, which perform $K$ aggregation steps on the input node features without any non-linearities and only a single, final linear map:
    \begin{equation}
        \bd{g} = \bd{\tilde{S}}^K\bd{X}\bd{w},
    \end{equation}
with weights $\bd{w} \in \R^{D \times 1}$ and latent representations $\bd{g} \in \R^{N\times 1}$.

Building on this model, we obtain the corresponding probabilistic formulation by placing a multivariate Gaussian prior on the weights $\bd{w}$.
Furthermore, we can transform the input signal with a feature map $\phi_\theta: \R \rightarrow \mathcal{H}$ that maps inputs to a potentially infinite-dimensional Hilbert space $\mathcal{H}$ and is parameterised by a set of hyperparameters $\theta$.
By subsequently marginalising the weights $\bd{w}$, we obtain an equivalent formulation 
    \begin{equation}
        \bd{g} = \bd{\tilde{S}}^K \bd{f},
    \end{equation}
where $\bd{f} \in \R^{N \times 1}$ is normally distributed with covariance matrix $[\bd{K}]_{ij} = \langle \phi_\theta(\bd{x}_i), \phi_\theta(\bd{x}_j) \rangle_\mathcal{H}$ and we assume $\bd{f}$ has zero mean.
The simplified graph convolution acts as a linear transformation on $\bd{f}$, hence the distribution of the resulting signal $\bd{g}$ is also Gaussian:
    \begin{equation}
        \bd{g} \sim \normal\left(\bd{0}, (\bd{\tilde{S}}^K) \bd{K} (\bd{\tilde{S}}^K)^T\right).
    \end{equation}
Thus, $\bd{g}$ corresponds to a Gaussian process on the domain whose structure is given by the graph $\mathcal{G}$. The covariance matrix $\bd{K}$ is computed by the \textit{node feature kernel} $k_\theta: \R^{D} \times \R^{D} \rightarrow \R$.

Going one step further, we take advantage of the ability of the Gaussian process to optimise hyperparameters to select between the number of graph convolutions to be applied. We achieve this by smoothly interpolating in each convolution step between the convolution matrix $\bd{\tilde{S}}$ and the identity matrix. The $k^{\text{th}}$ convolution matrix hence becomes $\bd{\tilde{S}}_k = \lambda_k \bd{\tilde{S}} + (1 - \lambda_k) \bd{I}$, where $\bd{\lambda} = [\lambda_1, \ldots, \lambda_K] \in [0; 1]^K$ are hyperparameters, subsequently referred to as the \textit{convolution weights}. The final Gaussian process prior thus becomes
    \begin{equation}
        \bd{g} \sim \normal\left(\bd{0}, (\bd{\tilde{S}}_1 \cdots \bd{\tilde{S}}_K) \bd{K} (\bd{\tilde{S}}_1^T \cdots \bd{\tilde{S}}_K^T)\right).\label{eq:gcgp}
    \end{equation}
A visualisation of the graph convolutional Gaussian process over nodes is shown in Figure~\ref{fig:model_summary} (b).

When fixing $K = 1$, $\lambda_1 = 1$ and using an asymmetric normalisation for the convolution matrix $\bd{\tilde{S}} = \bd{\tilde{D}}^{-1}\bd{\tilde{A}}$, we recover the Gaussian process model for semi-supervised node classification described by~\citealt{ng2018bayesian}.

\subsection{Model Interpretation}

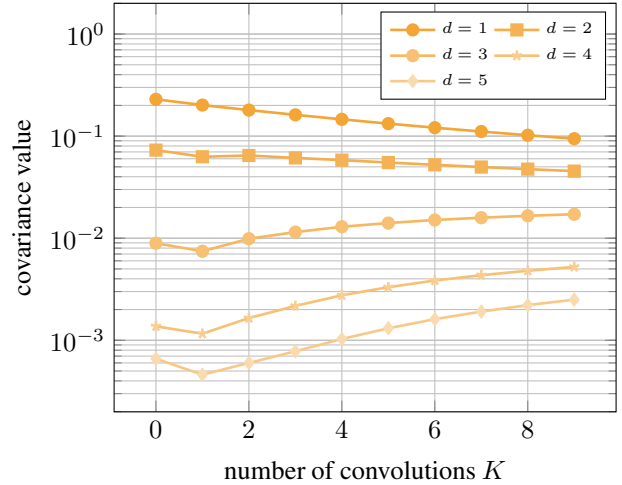
\begin{figure}
    \centering
    \begin{tikzpicture}
\begin{axis}[xlabel={number of convolutions $K$}, ylabel={covariance value}, ymin=0.0, ymax=2.0, ymode=log, width=\linewidth, height=0.85\linewidth, grid=both, legend style={font=\tiny, legend columns=2}]

\addplot+[highlightOrange, mark options={highlightOrange}, line width=1.0pt] table[x index=0,y index=1,col sep=comma] {covs.dat};
\addlegendentry{$d=1$}
\addplot+[highlightOrange!90, mark options={highlightOrange!85}, line width=1.0pt] table[x index=0,y index=2,col sep=comma] {covs.dat};
\addlegendentry{$d=2$}
\addplot+[highlightOrange!80, mark options={highlightOrange!70}, line width=1.0pt] table[x index=0,y index=3,col sep=comma] {covs.dat};
\addlegendentry{$d=3$}
\addplot+[highlightOrange!70, mark options={highlightOrange!55}, line width=1.0pt] table[x index=0,y index=4,col sep=comma] {covs.dat};
\addlegendentry{$d=4$}
\addplot+[highlightOrange!60, mark options={highlightOrange!40}, line width=1.0pt] table[x index=0,y index=5,col sep=comma] {covs.dat};
\addlegendentry{$d=5$}
\end{axis}
\end{tikzpicture}
\vspace{-0.25cm}
    \caption{Average covariances between nodes of varying geodesic distance. We randomly pick a node $i$ in the graph of the Yeast data set (for details, see Section~\ref{sec:datasets}). We then construct 5 disjoint sets of nodes that have geodesic distance of exactly $d=1, \ldots, 5$ from $i$ and compute the covariance between node $i$ and the nodes in each set, averaged over the nodes within the same set. We plot this mean covariance value for different number of convolutions $K$. We use an RBF-kernel as the node feature kernel with lengthscale and variance set to $1.0$.
    We find that as $K$ increases, the mean covariance values grow closer together, indicating that the covariance becomes spread more equally over the graph.}
    \label{fig:mean_covs}
\end{figure}

To obtain a better understanding of the effect of the simplified graph convolutions applied to the Gaussian process, we examine the prior covariance of $\bd{g}$ between two nodes for the case that all convolution weights have been set to $1$:
    \begin{equation}
        [\bd{\tilde{S}}^K\bd{K}\bd{\tilde{S}}^K]_{ij} = \sum_{\substack{k \in \mathcal{N}^K(i)\\ \cup \{i\}}} \sum_{\substack{l \in \mathcal{N}^K(j)\\ \cup \{j\}}} [\bd{\tilde{S}}^K]_{ik}[\bd{\tilde{S}}^K]_{lj} K_{kl}.\label{eq:kernel_elemwise}
    \end{equation}
Here, $\mathcal{N}^K(i)$ refers to the $K$-hop neighbourhood of node $i$.
We note that given the definition of the convolution matrix (cf. Equation~\ref{eq:agg_rule}), the coefficients $[\bd{\tilde{S}}^K]_{ik}$ and $[\bd{\tilde{S}}^K]_{lj}$ lie in the interval $[0, 1]$. Furthermore, for a fixed $K$, the $j^{\text{th}}$ entry of the $i^{\text{th}}$ row of $\bd{\tilde{S}}^K$ is non-zero if and only if $j$ is in the $K$-hop neighbourhood of $j$. Therefore, for larger $K$, more entries of $\bd{\tilde{S}}^K$ will be non-zero, as the size of the neighbourhood increases, yet every individual entry will be smaller because elements in $[0, 1]$ are being multiplied.
As a result, as we increase $K$, more but smaller terms are summed in Equation~\ref{eq:kernel_elemwise}, leading to the covariance to be spread across neighbourhoods of different sizes more equally. We confirm this empirically by choosing a random node in the input graph as the central node and plotting the average covariance of nodes at different geodesic distances. We observe that as $K$ is increased, the differences between the average covariance values start to shrink. The result is visualised in Figure~\ref{fig:mean_covs}.

\begin{figure}
    \centering
    \begin{tikzpicture}
\begin{axis}[xlabel={number of convolutions $K$}, ylabel={Dirichlet norm}, ymin=0, width=\linewidth, height=0.8\linewidth, grid=both]

\addplot+[highlightBlue, mark options={highlightBlue}, line width=1pt] table[x index=0,y index=1,col sep=comma] {dirichlet.dat};
\end{axis}
\end{tikzpicture}
\vspace{-0.25cm}
    \caption{Average Dirichlet norm of 5,000 functions sampled from the graph convolutional Gaussian process prior for varying number of convolutions $K$. We use the graph and \texttt{node2vec} features of the Yeast data set (for details, see Section~\ref{sec:datasets}) and an RBF-kernel as the node feature kernel. Its lengthscale and variance is set to $1.0$.
    For larger $K$, the average Dirichlet norm decreases, indicating that the sampled functions are smoother.}
    \label{fig:dirichlet_norms}
\end{figure}
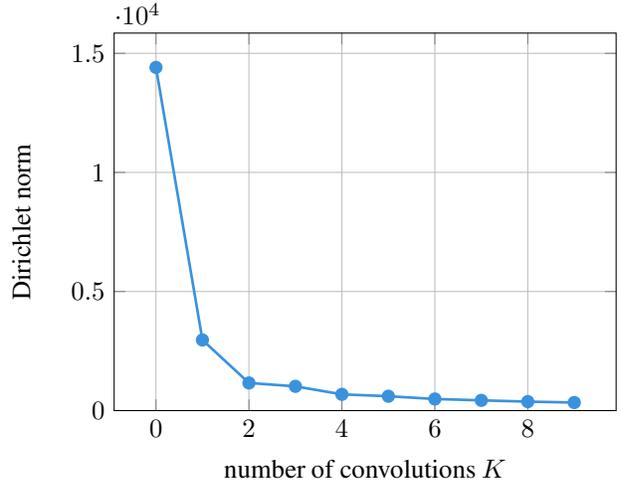

We expect a graph function for which the covariance between two distant nodes is higher to vary less from node to node compared to a function that has low covariance for distant nodes.
Hence, we expect the function to be smoother as measured by the Dirichlet norm
    \begin{equation}
        \left\lVert\bd{g}\right\rVert_\mathcal{G}^2 = \frac{1}{2} \sum_{i,j=1}^{N} a_{ij}(g_i - g_j)^2 = \bd{g}^T \bd{L} \bd{g},
    \end{equation}
where $\bd{L} = \bd{D} - \bd{A}$ is the graph Laplacian.
This agrees with the result of~\citealt{wu2019sgcn}, who have shown that simplified graph convolutions act as a low pass filter, thus smoothing the graph signal.
In Figure~\ref{fig:dirichlet_norms}, we plot the average Dirichlet norm for functions sampled from the described Gaussian process prior for varying $K$. As expected, the smoothness of the sampled functions increases for larger $K$.
We have found these observations to generalise well across data sets.

However, we note that higher smoothness does not necessarily result in better performance, which still depends on the labelled data. Yet, by optimising the ELBO with respect to the convolution weights $\bd{\lambda}$, we can adapt the smoothness of the posterior to fit the observed data.

\section{Sparse Variational Gaussian Processes for Link Prediction}\label{sec:link_pred}

The model described so far defines a Gaussian process over the nodes of the graph $\mathcal{G}$. In the following, we describe how to transform such a Gaussian process over nodes into a Gaussian process over node pairs to predict potential edges between them. We further introduce a variational inducing point approximation for the intractable posterior.

\subsection{Gaussian Processes over Pairs of Nodes}

A Gaussian process model over edges of an undirected graph must operate on the domain of pairs of nodes such that it is invariant to the order of the nodes within the pair. \citealt{yu2008link} propose to model edges using the function
    \begin{equation}
        r(\bd{x}_i, \bd{x}_j) = L^{-\frac{1}{2}} \sum_{l=1}^{L} g_l(\bd{x}_i)g_l(\bd{x}_j) - L^{\frac{1}{2}} k(\bd{x}_i, \bd{x}_j),\label{eq:edge_func}
    \end{equation}
where $\{g_l\}_{l=1}^{L}$ is a set of independent, identically distributed random variables with $g_l(\bd{x}) \sim \GP(\bd{0}, k(\bd{x}, \bd{x}'))$ modelling functions over the nodes of the graph.
In the limit of $L \rightarrow \infty$, $r$ converges to a Gaussian process over node pairs:
    \begin{equation}
        r(\bd{x}_i, \bd{x}_j) \sim \GP(\bd{0}, c((\bd{x}_i, \bd{x}_j), (\bd{x}_i', \bd{x}_j'))),
    \end{equation}
where $c((\bd{x}_i, \bd{x}_j), (\bd{x}_i', \bd{x}_j')) = k(\bd{x}_i, \bd{x}_i')k(\bd{x}_j, \bd{x}_j') + k(\bd{x}_i, \bd{x}_j')k(\bd{x}_j, \bd{x}_i')$~\citep[Theorem 2.2]{yu2008link}.
Crucially, the resulting Gaussian process has the desired property that its kernel $c$ is invariant to the order of the nodes within a pair.

As we would like the link prediction Gaussian process to incorporate neighbourhood information in its predictions, we define the set of random variables $\{g_l\}_{l=1}^{L}$ in Equation~\ref{eq:edge_func} to follow the graph convolutional Gaussian process prior defined in Equation~\ref{eq:gcgp}.
This results in the final graph convolutional Gaussian process model for link prediction:
    \begin{align}
        &\bd{f}(\bd{x}) \sim \GP(\bd{0}, \bd{K})\\[8pt]
        &\bd{g}(\bd{x}) \sim \GP(\bd{0}, \bd{\hat{K}} \equiv (\bd{\tilde{S}}_1 \cdots \bd{\tilde{S}}_K) \bd{K} (\bd{\tilde{S}}_1^T \cdots \bd{\tilde{S}}_K^T))\\[8pt]
        & \bd{r}(\bd{x}_i, \bd{x}_j) \sim \GP(\bd{0}, \bd{C}),\nonumber\\
        &\;\;\text{with } \bd{C}_{(i,j)(i',j')} = \bd{\hat{K}}_{ii'}\bd{\hat{K}}_{jj'} + \bd{\hat{K}}_{ij'}\bd{\hat{K}}_{ji'}.\label{eq:final_model}
    \end{align}
As before, $\bd{K}$ is computed using the node feature kernel $k_\theta: \R^{D \times 1} \times \R^{D \times 1} \rightarrow \R$ on the input node features. The full model is visualised in Figure~\ref{fig:model_summary}.

\subsection{Variational Inducing Point Approximation}

Predicting potential links between node pairs boils down to a binary classification problem, which dictates a Bernoulli likelihood. This leads to an intractable posterior distribution, which we will approximate with a variational distribution. We will also use a set of $M$ inducing points to reduce the computational complexity of inference (see Section~\ref{sec:complexity} for a detailed discussion).

Naively placing a set of of $M$ inducing points onto the signal domain $\R^{D \times 1}$, however, fails because of the functional form of the kernel $c$, which expects separate inputs for the two nodes of an edge. Hence, we require \textit{inducing edges} that are represented by pairs of inducing points.
We solve this problem by constructing a connected \textit{inducing graph} $\bar{\mathcal{G}} = (\bar{\mathcal{V}}, \bar{\mathcal{E}})$ with $\vert \bar{\mathcal{V}} \vert = \bar{N}$ and $\vert \bar{\mathcal{E}} \vert = \bar{E}$ and placing an inducing point $\bd{z}_i \in \R^{D \times 1}$ on each of the $\bar{N}$ nodes in the inducing graph. Each inducing point represents a node feature on the inducing graph.

In our experiments, we obtain the inducing graph by sampling from a modified Erd\H{o}s-R\'{e}nyi model~\cite{erdos1959random} with $\bar{E}$ edges that assures the sampled graph is connected.
We note that the exact structure of the inducing graph is of little importance and we have found that altering the type of random graph model does not affect the predictive performance. Unlike the input data, no graph convolutions are applied to the inducing points, hence the inducing graph merely forms the domain on which the inducing points lie. It is the inducing points, i.e. the node features of the inducing graph, which are optimised such that the inducing edges are most informative for posterior inference of missing links in the input graph.

The inducing points $\bd{z}_i \in \R^{D \times 1}$ on the nodes of the inducing graph are placed onto the domain of $\bd{f}$.
As the real input points, unlike the inducing points, are subject to a graph convolution, we have to employ inter-domain inference~\cite{lazaro2009inter, wilk2017cgp} for predicting missing links.
The inter-domain covariance between a node pair $(\bd{x}_i, \bd{x}_j)$ of the input graph and a node pair $(\bd{z}_{i'}, \bd{z}_{j'})$ of the inducing graph is given by
    \begin{align}
        \bd{C}_{(i,j)(i',j')} = &[(\bd{\tilde{S}}_1^T \cdots \bd{\tilde{S}}_K^T)\bd{K}_{\bd{XZ}}]_{ii'}[(\bd{\tilde{S}}_1^T \cdots \bd{\tilde{S}}_K^T)\bd{K}_{\bd{XZ}}]_{jj'}\nonumber\\ &+ [(\bd{\tilde{S}}_1^T \cdots \bd{\tilde{S}}_K^T)\bd{K}_{\bd{XZ}}]_{ij'}[(\bd{\tilde{S}}_1^T \cdots \bd{\tilde{S}}_K^T)\bd{K}_{\bd{XZ}}]_{ji'}.\nonumber
    \end{align}
$\bd{K}_{\bd{X}\bd{Z}}$ is computed using the node feature kernel $k_\theta: \R^{D \times 1} \times \R^{D \times 1} \rightarrow \R$ applied on node features of the input graph and the inducing graph. 

\subsection{Computational Complexity}\label{sec:complexity}

Using inducing points reduces the computational complexity of inference from $\LO(\vert\mathcal{E}\vert^3)$ for a vanilla Gaussian process to $\LO(\vert\mathcal{E}\vert\vert\bar{\mathcal{E}}\vert^2)$ with $\vert\bar{\mathcal{E}}\vert \ll \vert\mathcal{E}\vert$.
Furthermore, the variational lower bound allows us to optimise hyperparameters with stochastic gradient descent in mini-batches of $B$ edges each.
A potential computational bottleneck is computing the $B \times \vert\bar{\mathcal{E}}\vert$ covariance matrix of the graph-convolutional Gaussian process over nodes.
A naive implementation has complexity $\LO(\vert\mathcal{V}\vert^2\vert\bar{\mathcal{V}})$ due to the multiplication with the convolution matrix $\bd{\tilde{S}}$.
The computational complexity can be reduced by only considering the convolution matrix for the $N_{\text{pre-image}}$ nodes that lie in the pre-image of the convolution operation on the nodes incident to edges in the mini-batch.
This reduces computation complexity to $\LO(N_{\text{pre-image}}^2\vert\bar{\mathcal{V}})$.
$N_{\text{pre-image}}$ is typically small due to sparseness of most real-world graphs~\cite{goldenberg2010survey}.

\section{Experiments}\label{sec:experiments}

\subsection{Data Sets}\label{sec:datasets}

We apply our method on a set of benchmark data sets for link prediction as used by~\citealt{zhang2018link}, for example. The data sets are USAir~\cite{batagelj2006usair}, NS~\cite{newman2006ns}, PB~\cite{ackland2005pb}, Yeast~\cite{mering2002yeast}, C.ele~\cite{watts1998cele}, Power~\cite{watts1998cele}, Router~\cite{spring2004router}, and E.coli~\cite{zhang2018ecoli}. We provide an overview of the data set statistics in Table~\ref{tab:data_stats}.

The data sets come without node features, hence we generate them using the \texttt{node2vec} embedding generation algorithm introduced by~\citealt{grover2016node2vec} using the same setup as in~\citealt{zhang2018link}.
We use the typical way of computing the data set split into training and test set.
We randomly select $10\%$ of edges as test edges and remove them from the graph.
We then randomly select an equal number of node pairs that are not connected by an edge as negative test samples.
The remaining edges are used for the training set and we again select and equal number of pairs of unconnected nodes as negative training samples.
Note that we do not require a validation set as model selection will be performed based on the maximum ELBO achieved on the training set.

\begin{table}
    \centering
    \begin{tabular}{lrrr}
        \toprule
        \textbf{Data} & \textbf{\# nodes} & \textbf{\# edges} & \textbf{average node degree} \\
        \midrule
        USAir   & 332     & 2,126       & 12.81 \\
        NS      & 1,689   & 2,742       & 3.45 \\
        PB      & 1,222   & 16,714      & 27.36 \\
        Yeast   & 2,375   & 11,693      & 9.85 \\
        C.ele   & 297     & 2,148       & 14.46 \\
        Power   & 4,941   & 6,594       & 2.67 \\
        Router  & 5,022   & 6,258       & 2.49 \\
        E.coli  & 1,805   & 14,660      & 12.55 \\
        \bottomrule
    \end{tabular}
    \caption{Statistics of the data sets used in our experiments.}
    \label{tab:data_stats}
\end{table}

\subsection{Experimental Setup}
In all our experiments, we set the maximum number of convolutions to $K=2$.
We construct an inducing graph with $\vert \bar{V} \vert = \frac{\vert V \vert}{2}$ nodes and $\vert \bar{E} \vert = 2\vert \bar{V} \vert$ edges.
We use a radial basis function kernel (RBF) with automatic relevance determination (ARD)
    \begin{equation}
        k(\bd{x}, \bd{x'}) = \nu \exp\left(- \frac{1}{2} \sum\limits_{d=1}^{D} \frac{(x_d - x_d')^2}{l_d^2}\right)
    \end{equation}
for the base kernel of the graph convolutional Gaussian process. Its variance $\nu$ is initialised to $1.0$. Lengthscales $l_d$ for each feature $d$ are initialised to either $1.0$ or $2.0$ and the model with higher ELBO is selected.
The convolution weights $\lambda_1$ and $\lambda_2$ are initialised to $0.5$ and $0.3$ respectively.
The \texttt{node2vec} embeddings have size 128.
For parameter optimsation, we use the Adam optimiser~\cite{kingma2015adam} with a learning rate of $0.001$.
We train our models for up to 250 epochs and stop training early if the change in ELBO over $20$ epochs is less than $10^{-2}$. For all experiments we report the average performance and standard deviation over $5$ runs, each with different data splits. All experiments were performed on a NVIDIA Titan X GPU with Pascal architecture and 12GB of memory.

\subsection{Results}

\begin{table}
    \centering
    \resizebox{\linewidth}{!}{%
    \begin{tabular}{lccc}
        \toprule
        \textbf{Data} & \textbf{VGAE} & \textbf{LGP} & \textbf{GCLGP} \\
        \midrule
        USAir   & $89.28 \pm 1.99$          & $91.51 \pm 1.26$          & $\bd{95.01} \pm 0.65$ \\
        NS      & $\bd{94.04} \pm 1.64$     & $92.55 \pm 1.89$          & $92.47 \pm 1.60$ \\
        PB      & $\bd{90.70} \pm 0.53$     & $87.99 \pm 0.19$          & $90.33 \pm 0.53$ \\
        Yeast   & $93.88 \pm 0.21$          & $95.02 \pm 0.40$          & $\bd{95.83} \pm 0.27$ \\
        C.ele   & $81.80 \pm 2.18$          & $80.35 \pm 1.37$          & $\bd{84.27} \pm 1.99$ \\
        Power   & $71.20 \pm 1.65$          & $76.10 \pm 0.75$          & $\bd{79.30} \pm 1.81$ \\
        Router  & $61.51 \pm 1.22$          & $69.61 \pm 1.33$          & $\bd{79.79} \pm 3.62$ \\
        E.coli  & $90.81 \pm 0.63$          & $\bd{94.25} \pm 0.19$     & $93.89 \pm 0.62$ \\
        \bottomrule
    \end{tabular}}
    \caption{The proposed graph-convolutional Gaussian process (Equation~\ref{eq:final_model}) compared to the non-convolutional Gaussian process with the kernel from~\citealt{yu2008link} and the variational graph auto-encoder~\cite{kipf2016gvae} in terms of area under the ROC curve (\textbf{AUC}). Results reported with one standard deviation.}
    \label{fig:results_auc}
\end{table}

\begin{table}
    \centering
    \resizebox{\linewidth}{!}{%
    \begin{tabular}{lccc}
        \toprule
        \textbf{Data} & \textbf{VGAE} & \textbf{LGP} & \textbf{GCLGP} \\
        \midrule
        USAir   & $89.27 \pm 1.29$          & $82.49 \pm 1.58$          & $\bd{89.82} \pm 1.30$ \\
        NS      & $\bd{95.83} \pm 1.04$     & $92.66 \pm 2.44$          & $90.31 \pm 3.89$ \\
        PB      & $\bd{90.38} \pm 0.72$     & $77.77 \pm 4.64$          & $84.66 \pm 0.46$ \\
        Yeast   & $\bd{95.19} \pm 0.38$     & $92.07 \pm 0.95$          & $94.52 \pm 0.39$ \\
        C.ele   & $\bd{78.32} \pm 3.49$     & $71.88 \pm 1.27$          & $77.35 \pm 2.11$ \\
        Power   & $75.91 \pm 1.56$          & $\bd{88.86} \pm 3.11$     & $86.83 \pm 1.74$ \\
        Router  & $70.36 \pm 0.85$          & $79.82 \pm 4.01$          & $\bd{91.23} \pm 0.74$ \\
        E.coli  & $92.77 \pm 0.65$          & $93.99 \pm 0.47$          & $\bd{95.02} \pm 1.08$ \\
        \bottomrule
    \end{tabular}}
    \caption{The proposed graph-convolutional Gaussian process (Equation~\ref{eq:final_model}) compared to the non-convolutional Gaussian process with the kernel from~\citealt{yu2008link} and the variational graph auto-encoder~\cite{kipf2016gvae} in terms of average precision (\textbf{AP}). Results reported with one standard deviation.}
    \label{fig:results_ap}
\end{table}

We compare the graph convolutional Gaussian process model for link prediction (GCLGP) from Equation~\ref{eq:final_model} to a non-convolutional Gaussian process with the kernel proposed by~\citealt{yu2008link} (LGP). Moreover, we compare it to the variational graph auto-encoder (VGAE) described by~\citealt{kipf2016gvae}, which is also a probabilistic model and hence most closely related to our approach. We compare models in terms of area under the receiver-operating characteristic curve (AUC) and average precision (AP).

The results of our experiments are presented in Table~\ref{fig:results_auc} (AUC) and Table~\ref{fig:results_ap} (AP).
When comparing the proposed GCLGP to the non-convolutional LGP, we find that the former outperforms the latter on most data sets, by up to $10.0$ in terms of AUC for some data sets. We find improved performance on six out of the eight data sets in terms of AUC and also six out of eight in terms of AP. In other cases, such as the NS data set, LGP performs only marginally better in terms of AUC, relative to the standard deviation.
The results show that GCLGP is able to use local neighbourhood information and that this is indeed beneficial for most of the data sets we are evaluating on.
Our method outperforms VGAE in terms of AUC on six out of the eight data sets, often by a large margin (by over $15.0$ on the Router data set). 
GCLGP is roughly on par with VGAE in terms of AP, outperforming it on four of the eight data sets. The results highlight the strengths of this highly flexible probabilistic modelling approach.

\section{Conclusion}

We have described a Gaussian process model for link prediction incorporating both node features and local neighbourhood information.
We have introduced a graph convolutional Gaussian process over nodes based on a kernel that allows to interpolate between node neighbourhoods of different sizes.
We have shown how the model is applied to the task of link prediction and introduced a variational inducing point method for Gaussian processes over pairs of nodes that places inducing points on the nodes of a randomly generated, connected inducing graph.
Finally, we have shown that the proposed model exhibits strong performance on a range of graph data sets, outperforming non-convolutional Gaussian processes and the graph neural network-based variational autoencoder in many settings.
Future work can investigate the benefits of deeper hierarchies of Gaussian processes for link prediction.

\section*{Acknowledgements}

FLO acknowledges funding from the Huawei Studentship at the Department of Computer Science and Technology of the University of Cambridge.
We would like to thank C\u{a}t\u{a}lina Cangea, Cristian Bodnar, Alessandro Di Stefano, Duo Wang, Simeon Spasov and Ramon Vi\~{n}as for helpful discussions and feedback.


\bibliography{main}
\bibliographystyle{icml2020}





\end{document}